\documentclass{article} 
\usepackage{nips15submit_e,times}
\usepackage{hyperref}
\usepackage{url}

\usepackage{amsmath,amssymb}
\usepackage{algorithm}
\usepackage{algpseudocode}

\usepackage{graphicx}
\usepackage{hyperref}
\usepackage{subfigure} 
\usepackage{wrapfig}
\usepackage{mwe} 
\usepackage{multicol}

\usepackage{mathtools}

\nipsfinalcopy 
\title{Next Generation Multicuts for Semi-Planar Graphs}

\author{
Julian Yarkony  \\ 
Experian Data Lab
}

\begin{document}
\maketitle
\begin{abstract}
We study the problem of multicut segmentation.  We introduce modified versions of the Semi-PlanarCC based on bounding Lagrange multipliers.  We apply our work to natural image segmentation.  
\end{abstract}
\section{Introduction}
In this work, we formulate image segmentation from the perspective of constructing a multicut \cite{deza1997geometry} over the set of image pixels/superpixels \cite{superPx} that agrees closely with an input set of noisy pairwise affinities.  A multicut is a partition of a graph into an arbitrary number of connected components.  However the number of components is not a user defined hyper-parameter that must be hand tuned but arises naturally as a function of the noisy pairwise affinities.  Finding the optimal multicut is NP-hard even on a planar graph \cite{nphardplanar} however there has been much successful work concerning the application of multicuts despite this fundamental difficulty \cite{highcc,bagon,Andres2012}.  We focus on the case where the input affinities are specified by a planar graph augmented with a specific class of long range interactions which  \cite{AndresYarkony2013} calls repulsive.  These interactions create hard constraints that require that given pairs of non-adjacent nodes be in separate components. 

%
%
  Consider a planar graph $G$ that is defined by a set of edges $E$ and a set of nodes $V$.  We index $E$ using $e$ or $\hat{e}$.  We  refer to the nodes connected  by an edge $e$ as $e_1$ and $e_2$ respectively.  We define a partitioning (also called a multicut) of $G$ into an arbitrary number of components using a vector $\bar{X} \in \{0,1 \}^{|E|}$ and index $\bar{X}$ using $e$.  We use $\bar{X}_e=1$ to indicate that the two nodes $e_1$ and $e_2$ are in separate components and use $\bar{X}_e=0$ otherwise.  For short hand we refer to  an edge $e$ as ``cut" if $e_1$ and $e_2$ are in separate components and otherwise  refer to $e$ as ``uncut".  

We denote the set of $\bar{X}$ that define a multicut using $MCUT$.   We define $MCUT$ using the following notation.  Let $C$ denote the set of  all cycles in $G$ and index $C$ with $c$.  An element $c\in C$ is itself a set where its elements are the edges on the cycle $c$.  We use $c-\hat{e}$ to refer to the set of edges in cycle $c$ excluding edge $\hat{e}$. 
We express $MCUT$ in terms of constraints as follows \cite{ilpalg}\cite{Andres2012}.  For every cut edge $\hat{e}$ and for every cycle $c$ containing $\hat{e}$ there must be a cut edge in addition to $\hat{e}$ on the cycle.  Constraints of this form are called ``cycle inequalities".  We formally describe $MCUT$ below.   $MCUT = \{ \bar{X} \in \{ 0,1\}^{|E|}: \; \; \; \sum_{e \in c-\hat{e}}\bar{X}_e\geq \bar{X}_{\hat{e}} \; \; \forall c \in C, \hat{e} \in c \}$

The LP relaxation of $MCUT$ to the range [0,1] is denoted $CYC$.  Multicuts are associated with a cost using a real valued problem instance specific vector $\theta \in \mathcal{R}^{|E|}$.  We use the dot product of $\theta$ and $X$ or $\bar{X}$ to define the cost of a multicut $X$ or $\bar{X}$.  The objectives that are minimized over $MCUT$ and $CYC$ respectively are  $\min_{\bar{X}\in MCUT} \theta \bar{X}$ and  $\min_{X\in CYC} \theta X$ . 




\section{An Alternative Representation of Multicut:   The Cut Cone}

We  represent multicuts in a different manner using the method of \cite{HPlanarCC} which we now discuss. We refer to the set of 2-colorable multicuts  of $G$ as $C_2$ and index it with $r$.  We use a matrix $Z$ to represent $C_2$ where $Z\in \{ 0,1\}^{|E|\times |C_2|}$.  We use $Z_{er}=1$ to indicate that edge $e$ is cut in multicut $r$.    Similarly $Z_{er}=0$ indicates that edge $e$ is not cut in multicut $r$.  We define $\gamma$ to be a non-negative real valued vector with cardinality $|C_2|$ which we use to express a multicut. We use  $(Z\gamma)_e$ to refer to the value of $Z\gamma$ for edge $e$.  In \cite{HPlanarCC} the following properties of $Z\gamma$ are established. 


\begin{align}
\label{mcutded}
MCUT = \{ Z\gamma ; \; \; \gamma \geq 0, (Z\gamma)_e \in \{ 0,1\} \quad \forall e \} \quad \quad
CYC = \{ Z\gamma ; \; \; \gamma \geq 0, (Z\gamma)_e \in [ 0,1] \quad \forall e \} 
\end{align}

The cut cone is associated with an extension of the cut cone called the expanded cut cone \cite{HPlanarCC} which only restricts $\gamma$ to be non-negative.  The expanded cut cone is associated with a slack term $\beta$ which is used to provide a penalty for cutting edges more than once.  That penalty is $-\min(0,\theta)$, denoted $-\theta^-$, is of cardinality $|E|$ and is indexed by $e$.  Here $\theta^-_e=\min(0,\theta_e)$. We write optimization over the expanded cut cone below.  

\begin{align}
\min_{\substack{\gamma \geq 0\\ \beta\geq 0}} \theta Z \gamma -\theta^- \beta \quad s.t. \quad Z\gamma-1 \leq \beta
\end{align}

 It is established that the introduction of $\beta$ does not alter the value of the LP relaxation \cite{HPlanarCC}.  We use $\min(1,Z\gamma)$ to refer to the element-wise minimum of $1$ and $Z\gamma$.  In \cite{HPlanarCC} it is established that given any solution $\gamma$ that $X=\min(Z\gamma,1)$ lies in $CYC$,  Furthermore if $X=\min(Z\gamma,1) \in \{ 0,1\}^{|E|}$ then $X \in MCUT$.   The expanded cut cone is useful in the setting in which we are unable to access the full set of columns of $Z$.  This additional flexibility allows for the application of efficient delayed column generation algorithms \cite{HPlanarCC,PlanarCC}.  


We now present our novel contribution.   Consider a set of pairs of nodes in $G$ denoted $F$ which is indexed by $f$.  The nodes on pair $f$ are denoted $f_1$ and $f_2$ and need not be adjacent in $G$.  We require that for each $f$ that $f_1$ and $f_2$ are in separate components in any multicut.    
We refer to the set of  all paths  between pairs in $F$ as $P$ and index it with $p$.  Here each $p$ is  defined by a connected path of edges in $G$.  
 Each path $p$ is associated with a pair $f$ in $F$ and the first and last nodes on the path are $f_1$ and $f_2$ respectively. 
We now write a constrained optimization over the expanded cut cone to enforce that pairs in $F$ are in separate components.    

\begin{align}
\min_{\substack{\gamma \geq 0\\ \beta\geq 0}} \theta Z \gamma -\theta \beta \quad s.t. \quad  Z\gamma-1 \leq \beta \quad \quad  \sum_{e \in p} (Z\gamma)_e \geq 1 \quad \forall p \in P
\end{align}

Inspired by the slack terms used in \cite{HPlanarCC} which the authors denote $\alpha$ we introduce a set of slack terms to assist in satisfying $\sum_{e \in p} (Z\gamma)_e \geq 1$.  We allow any edge to be cut more than as described in $Z\gamma$ but  at a cost.  We denote the cost vector as  $\theta^+ \in \mathcal {R}^{|E|}_{0+}$ where $\theta^+_e=\max(0,\theta_e)$.  We indicate the slack term for cutting  edges as  $\kappa \in \mathcal{R}^{|E|}_{0+}$.  For notational ease we introduce a matrix $S \in \{ 0,1\}^{|P| \times |E|}$.  Here $S$ has one row for each path $p \in P$ and is indexed by $pe$.  Here $S_{pe}=1$ if and only if edge $e$ is on the path $p$.  We write the modified LP below.  
 

\begin{align}
\label{lpfinmat}
\min_{\substack{\gamma \geq 0, \beta\geq 0 , \kappa \geq 0}} \theta Z \gamma -\theta \beta +\theta^+ \kappa \quad ; \quad Z\gamma-1 \leq \beta \quad ; \quad SZ\gamma+ S\kappa \geq 1
\end{align}

We use $X=\min(1,Z\gamma+\kappa)$ to denote the element wise minimum of $[1, Z\gamma+\kappa]$. In Section \ref{proofcut} we show that $X=\min(1,Z\gamma+\kappa) \in CYC$ for all optimizing solutions to the LP in Eq \ref{lpfinmat}. 

\section{Dual Formulation}


Solving the primal problem in Eq \ref{lpfinmat} is challenging because of the exponential number of primal constraints and primal variables.  Past work has employed cutting plane methods in the dual.  Thus we analyze the dual form of this objective which we write below using $T$ to denote transpose.  
\begin{align}
\label{DualLpFinProperA}
 \max_{\substack{\lambda \geq 0 \\ \psi \geq 0}}- 1^T \lambda^T +1^T\psi^T  \quad s.t. \quad 
  Z^T\theta^T \geq Z^T S^T\psi^T -Z^T\lambda^T \quad ;\quad 
 \theta^{+T} \geq  S^T\psi^T \quad ;\quad 
 -\theta^{-T}\geq \lambda^T 
\end{align}  
We rely on a column generation and cutting plane methods jointly to construct a sufficient subset of the columns of  $S$ and $Z$ so as to fully optimize the  objective.  We denote these subsets as $\hat{S}$ and $\hat{Z}$.  Observe that the introduction of $\beta$ and $\kappa$ results in bounds on the Lagrange multipliers $\lambda$ and $\psi$.  We observe in experiments that without these bounds the optimization fails to converge on most problems.  

Observe that $ Z^T\theta^T \geq Z^T S^T\psi^T -Z^T\lambda^T$ is defined on the planar graph $G$.   We write the most violated constraint of that form as the solution as to the following objective:  $\min_{z \in C_2}(\lambda+\theta -\psi \hat{S})z$.  Finding the lowest cost 2-colorable multicut of a planar graph is solvable in $O(|E|^{3/2}\log |E|)$ time \cite{maxcutuni,PlanarCC,bar1,bar2,bar3,fisher2}. In practice after computing $z$, all cuts isolating a  component are added to $\hat{Z}$.

Naively finding violated primal constraints can be done by studying the solution to the primal problem.  First we obtain the primal solution from the LP solver after solving the dual.  We then find violated paths in the primal solution $X=\min(1,\hat{Z}\gamma +\kappa)$ via shortest path calculation \cite{ilpalg} between nodes in each pair $f\in F$. 

\subsection{Finding Better Primal Constraints via Path Pursuit}
Inspired by the cycle pursuit approach of \cite{sontag,sontag2} we introduce an approach that we call ``path pursuit" designed to find paths that increase the dual objective as quickly as possible.   
We apply the approximation that all paths cross a 2-colorable multicut cut no more than once.  We now consider a set of slack variables $\nu$ of cardinality $|E|$ which is indexed by $e$ and defined as follows.
\begin{align}
\nu_e=\min[\theta^+_e-(S^T \psi^T)_e,\min_{r;\; s.t. \; \hat{Z}_{er}=1}(\hat{Z}_{:r})^T\theta^T - (\hat{Z}_{:r})^T S^T\psi^T +(\hat{Z}_{:r})^T\lambda^T]
\end{align} 
Consider any path $p$.  We can increase $\psi_p$ greedily by the smallest $\nu_e$ on the path without violating identified constraints.  Finding the path that maximizes the minimum slack is called the widest path problem which solvable in time equal to shortest path calculation. Once this path is identified we set $\psi_p$ to the maximum value allowed given that all other dual variables are fixed.  If no  path of non zero width exists then we use the naive approach.  We now formalize our approach in Algorithm \ref{dualsolvesimple}.

\begin{algorithm}
\caption{Dual Optimization }
\begin{algorithmic} 

\State $\hat{S} \leftarrow \{ \}$,  $\hat{Z}\leftarrow \{ \}$\\
\While{Violated constraints exist}
\State $[\lambda,\psi,\gamma,\kappa] \leftarrow$ Solve Eq \ref{lpfinmat}, \ref{DualLpFinProperA} given $\hat{Z},\hat{S}$\\
\State $z \leftarrow \min_{z \in C_2 }(\theta+\lambda+\psi \hat{S})z$ \quad $ \{ z_1,z_2\ldots \} \leftarrow isocuts(z)$\quad $ \hat{Z }\leftarrow \hat{Z} \cup \{ z_1,z_2\ldots \}$ 
\For{$f \in F$}
\If{ there exists a violated path for pair $f_1$,$f_2$}
	\State Compute a row $p$  and add it to $\hat{S}$  for pair $f$ using either widest and/or shortest path calculation. Widest path calculation allows for multiple rows to be computed
\EndIf
\EndFor
 \EndWhile
\end{algorithmic}
  \label{dualsolvesimple}
\end{algorithm}

If widest path computation is used we refer to the algorithm as Alg 1 and otherwise as Alg 2.
\subsection{Computing Upper and Lower Bounds}
At any time one can identify an upper bound on the optimal multicut.  First one produces an $X=\min(\hat{Z}\gamma +\kappa)$.  This is obtained for ``free" because the CPLEX LP solver provides the primal solution $\gamma,\beta,\kappa$ whenever it solves for $\lambda,\psi$.  Next for unique value $\mu$ in $X$ set $\bar{X}\leftarrow X\geq \mu$.  Then uncut all edges in the middle of a  component.  Next check for all $f\in F$  that $f_1,f_2$ are in separate components.  If this is satisfied compute the value  $\theta \bar{X}$ and retain the solution $\bar{X}$ if it is the lowest value solution computed so far.  In practice we only use a few values of $\mu$ such as $\{ 0.2, 0.4, 0.6,0.8\} $ 

In Section \ref{proofbound} we establish that at any time a lower bound on the optimal integer solution is equal to the value of the LP plus $\frac{3}{2}$ times the the value of $\min_{z \in C_2 }(\theta+\lambda+\psi \hat{S})z$.  
\section{Experiments}

We evaluate our two algorithms on the Berkeley segmentation data set (BSDS)\cite{bsdspaper}.  For each image in the BSDS test set we are provided with superpixels, and potentials $\theta$ defined  between adjacent superpixels by the authors of \cite{PlanarCC}. For each image we then select a number of random pairs of superpixels that are in separate components according to at least two ground truth annotators.  We constructed examples with $[28,58,208,308,408,508]$ pairs. 
For a baseline we employed the original Semi-PlanarCC algorithm of \cite{AndresYarkony2013} but it does not converge on most  of our problems so we did not plot it.  This is interesting because the major difference between our algorithm with naive addition of constraints (not widest path calculation) is simply the introduction of $\kappa$ which thus establishes the value of the $\kappa$ term.  

\begin{figure*}
\centering     
\subfigure{\label{quantprb:a}\includegraphics[width=30mm]{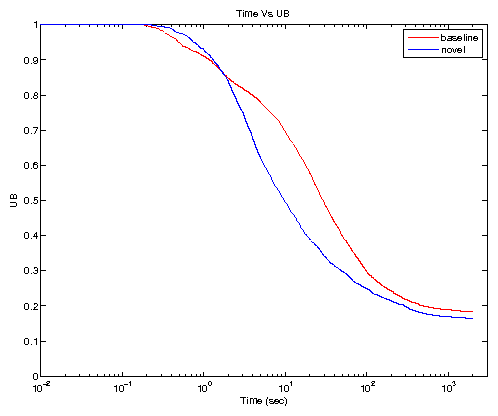}}
\subfigure{\label{quantprb:a}\includegraphics[width=30mm]{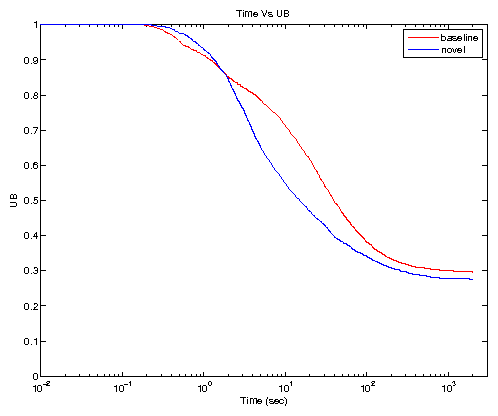}}
\subfigure{\label{quantprb:a}\includegraphics[width=30mm]{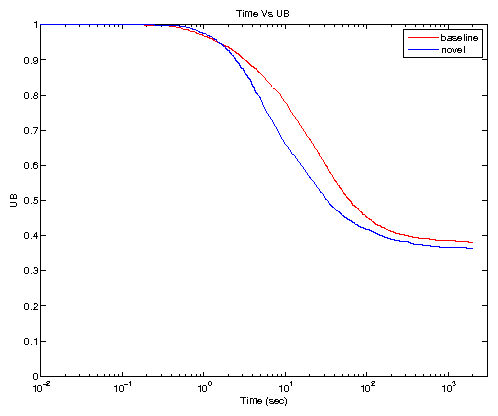}}
\subfigure{\label{quantprb:a}\includegraphics[width=10mm]{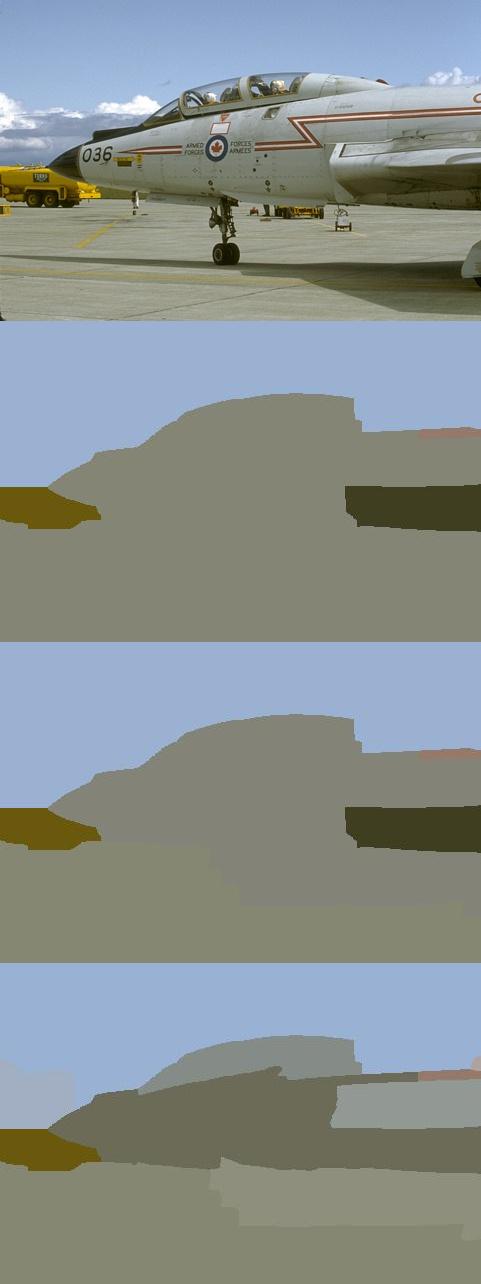}}
\subfigure{\label{quantprb:a}\includegraphics[width=10mm]{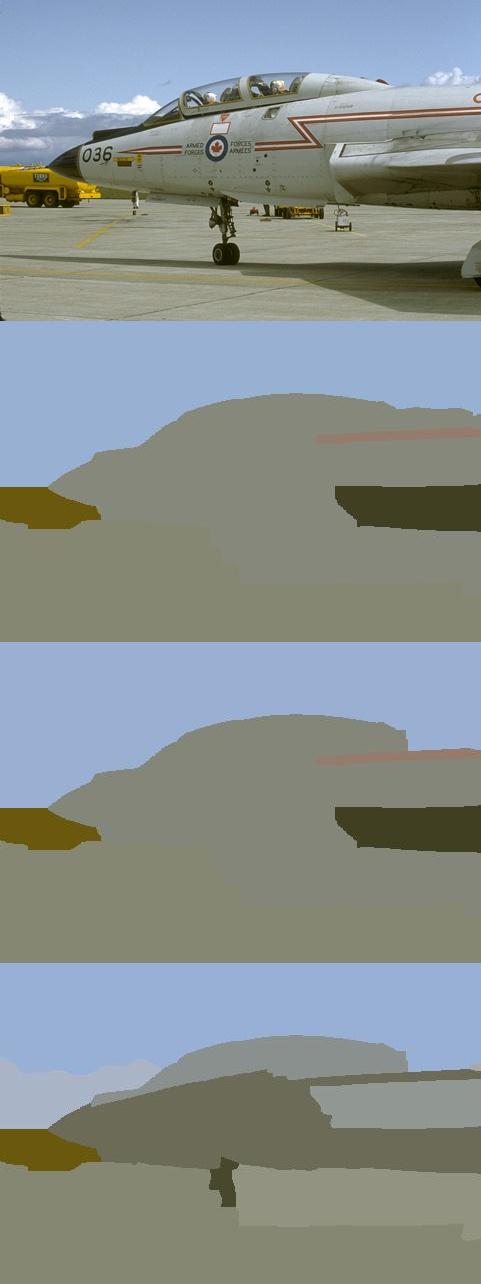}}
\subfigure{\label{quantprb:a}\includegraphics[width=10mm]{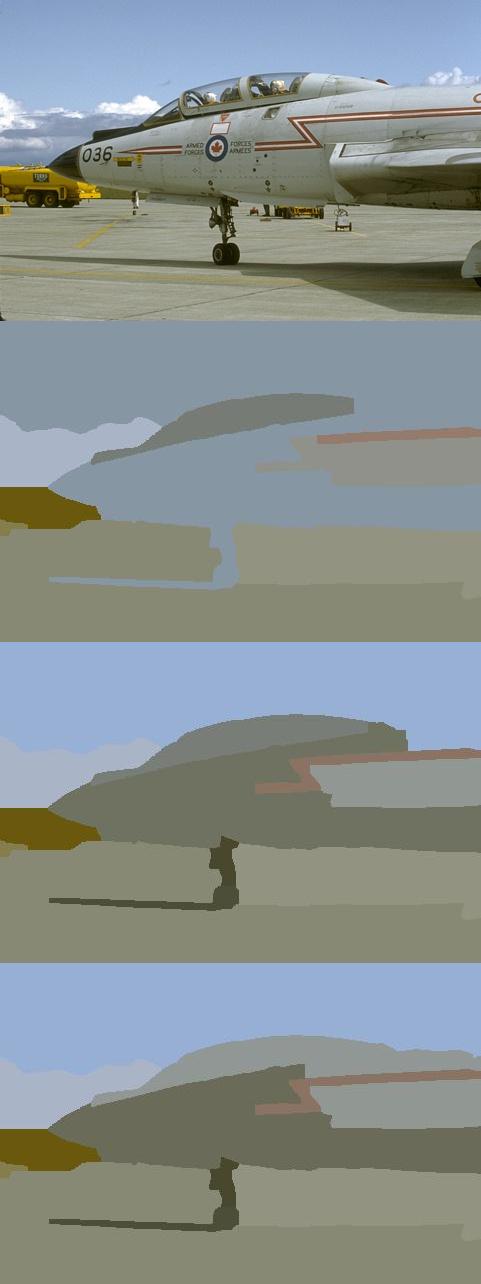}}\\
\subfigure{\label{quantprb:a}\includegraphics[width=10mm]{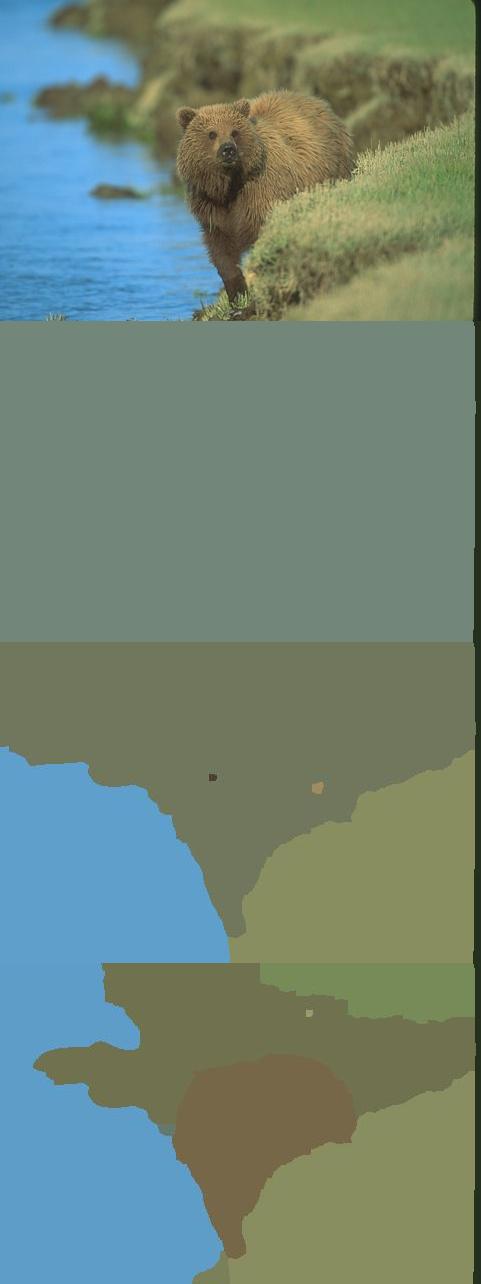}}
\subfigure{\label{quantprb:a}\includegraphics[width=10mm]{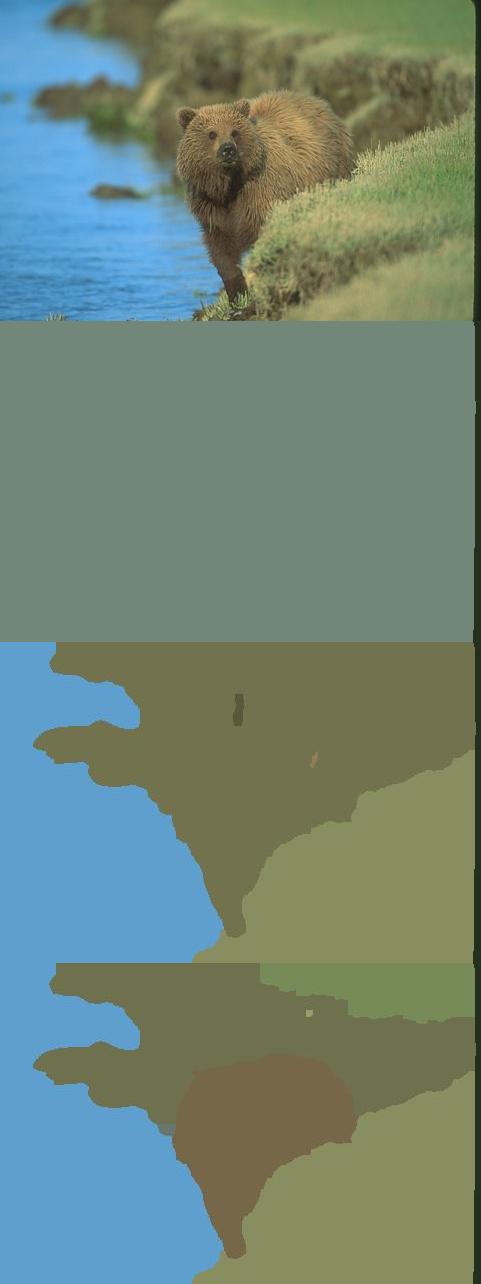}}
\subfigure{\label{quantprb:a}\includegraphics[width=10mm]{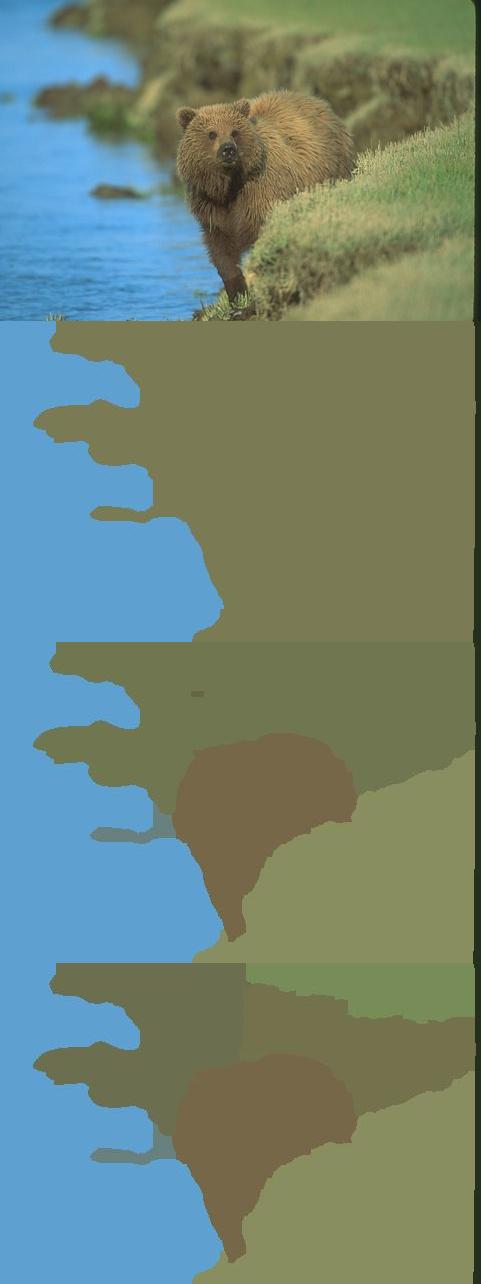}}
\subfigure{\label{quantprb:a}\includegraphics[width=10mm]{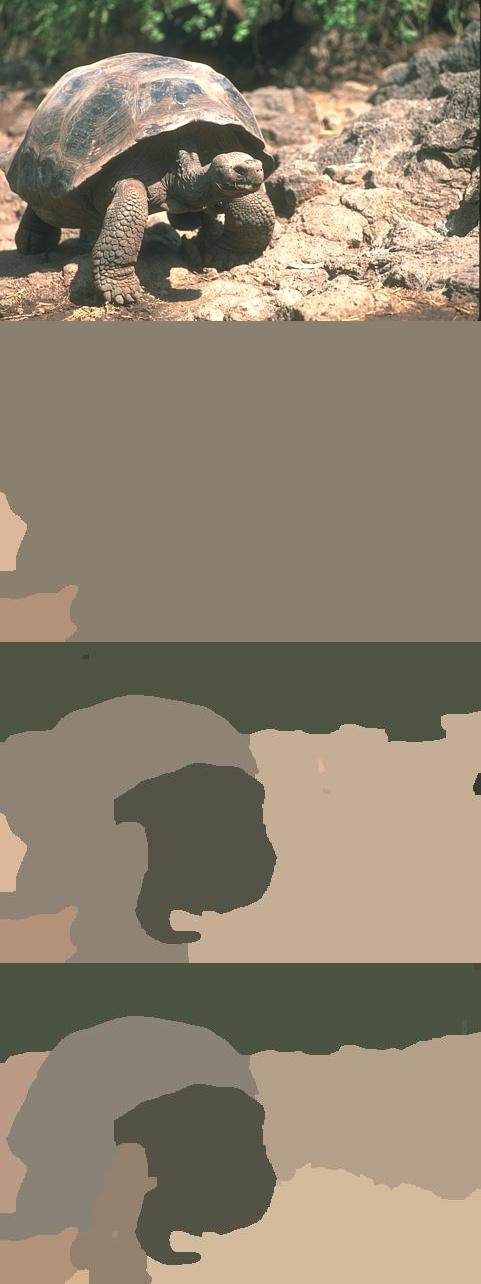}}
\subfigure{\label{quantprb:a}\includegraphics[width=10mm]{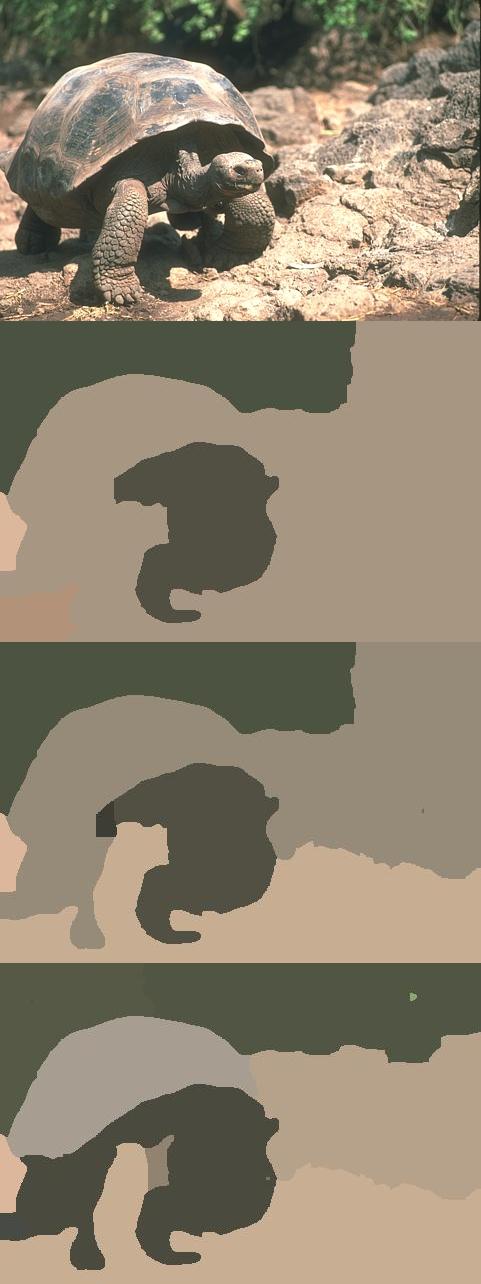}}
\subfigure{\label{quantprb:a}\includegraphics[width=10mm]{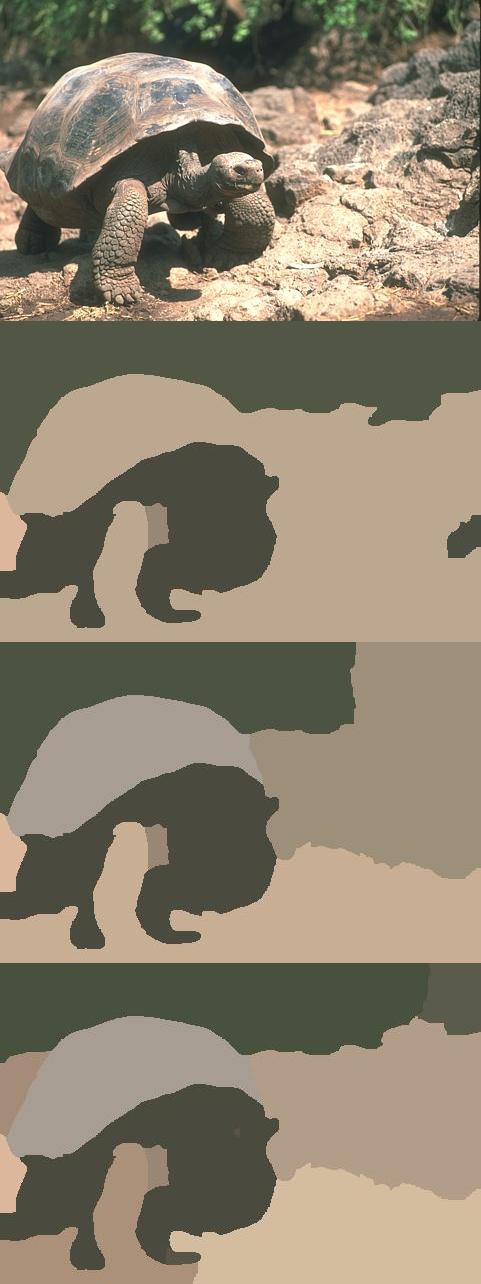}}
\subfigure{\label{quantprb:a}\includegraphics[width=5mm]{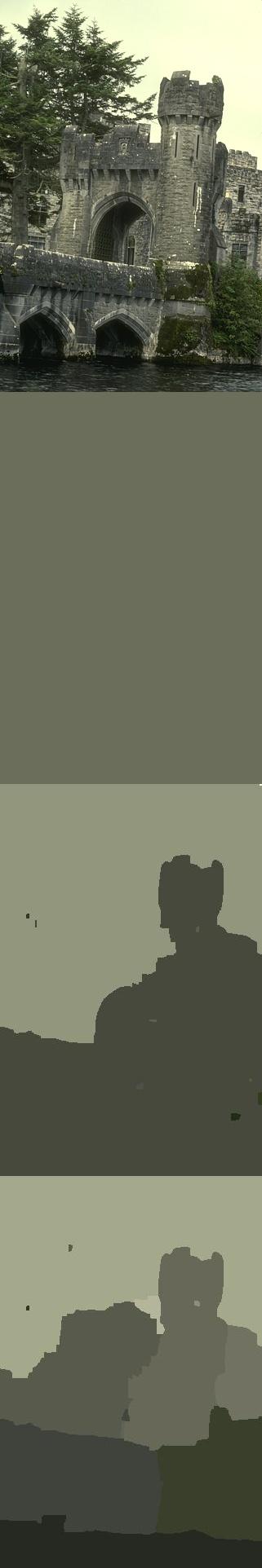}}
\subfigure{\label{quantprb:a}\includegraphics[width=5mm]{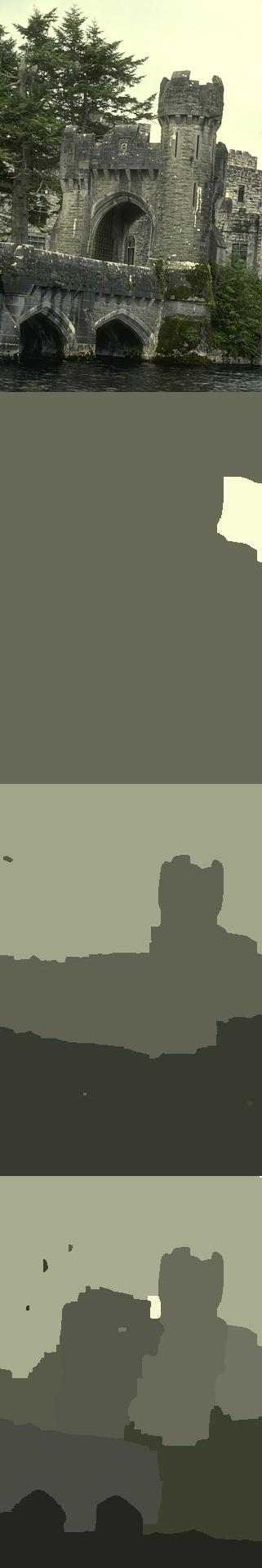}}
\subfigure{\label{quantprb:a}\includegraphics[width=5mm]{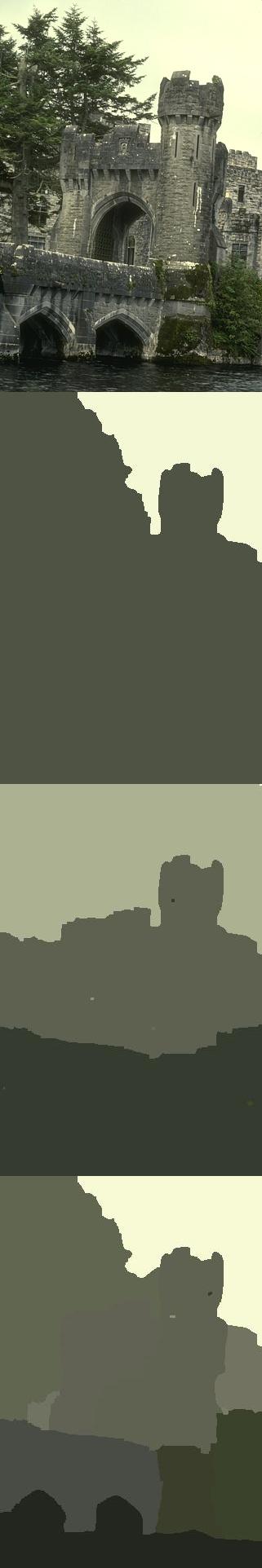}}
\subfigure{\label{quantprb:a}\includegraphics[width=5mm]{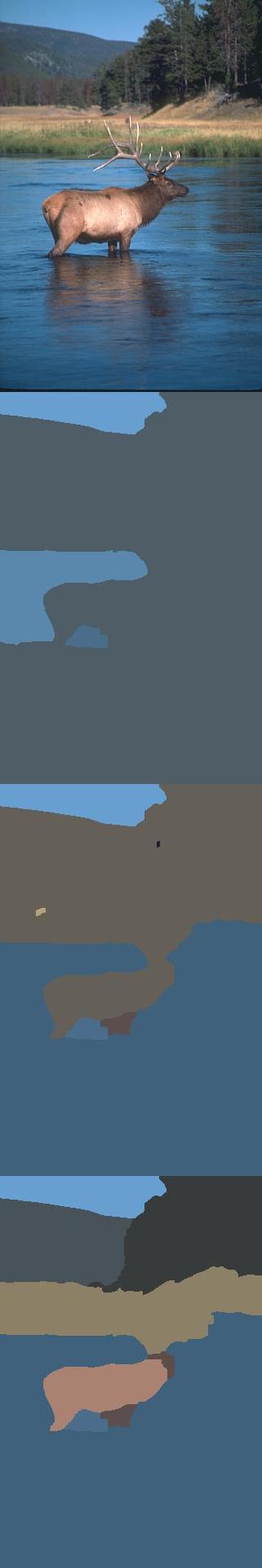}}
\subfigure{\label{quantprb:a}\includegraphics[width=5mm]{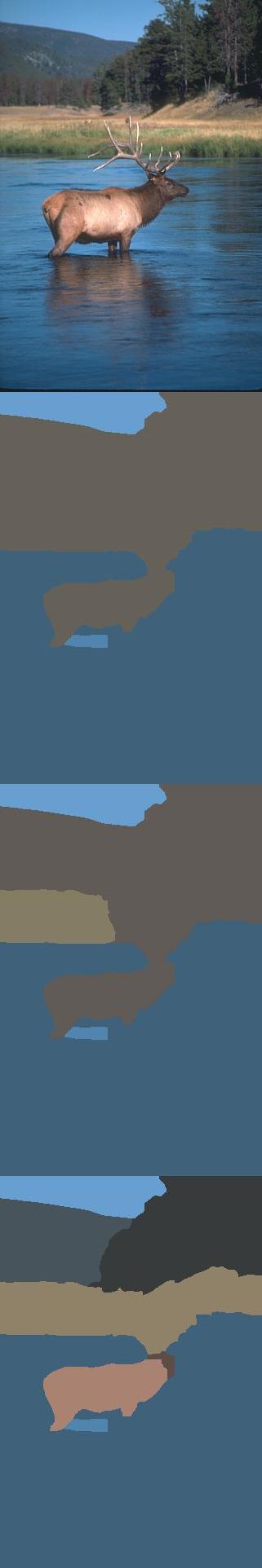}}
\subfigure{\label{quantprb:a}\includegraphics[width=5mm]{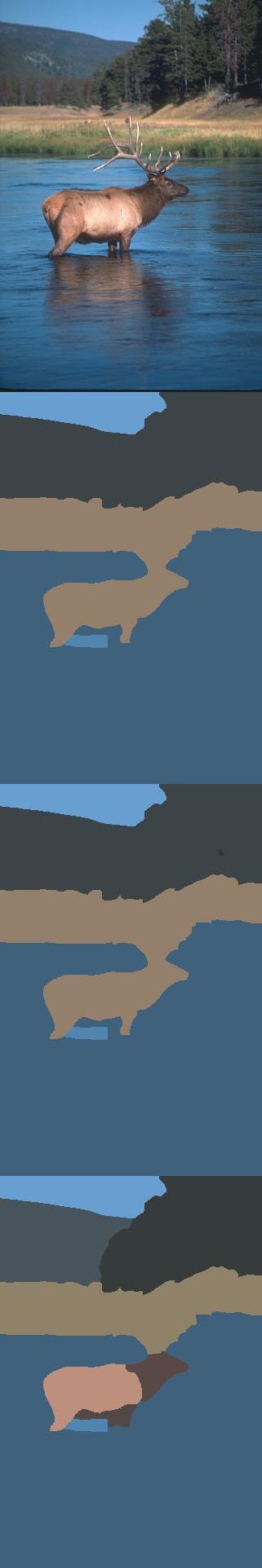}}
\subfigure{\label{quantprb:a}\includegraphics[width=10mm]{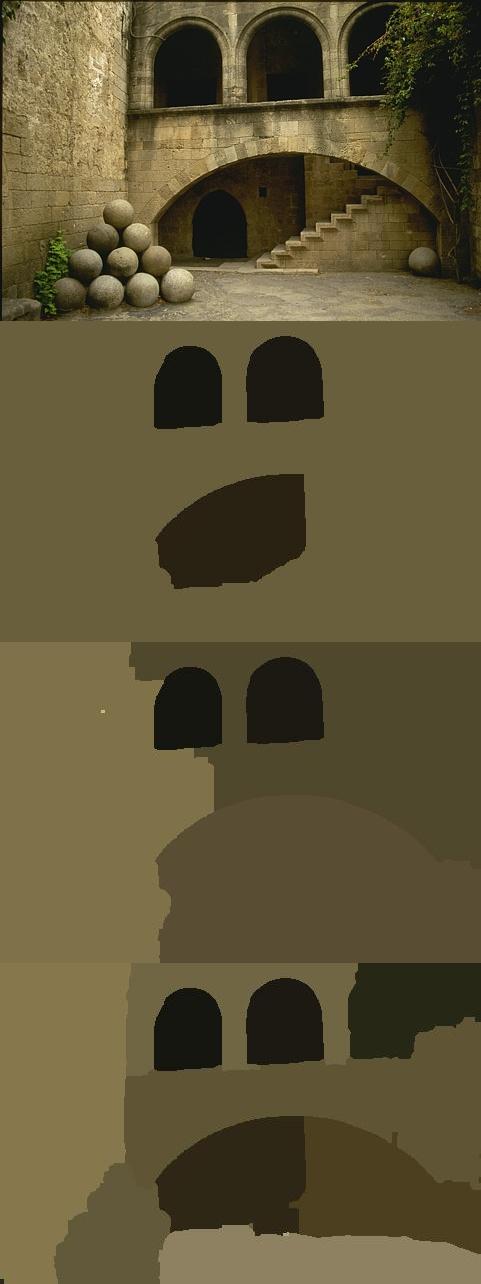}}
\subfigure{\label{quantprb:a}\includegraphics[width=10mm]{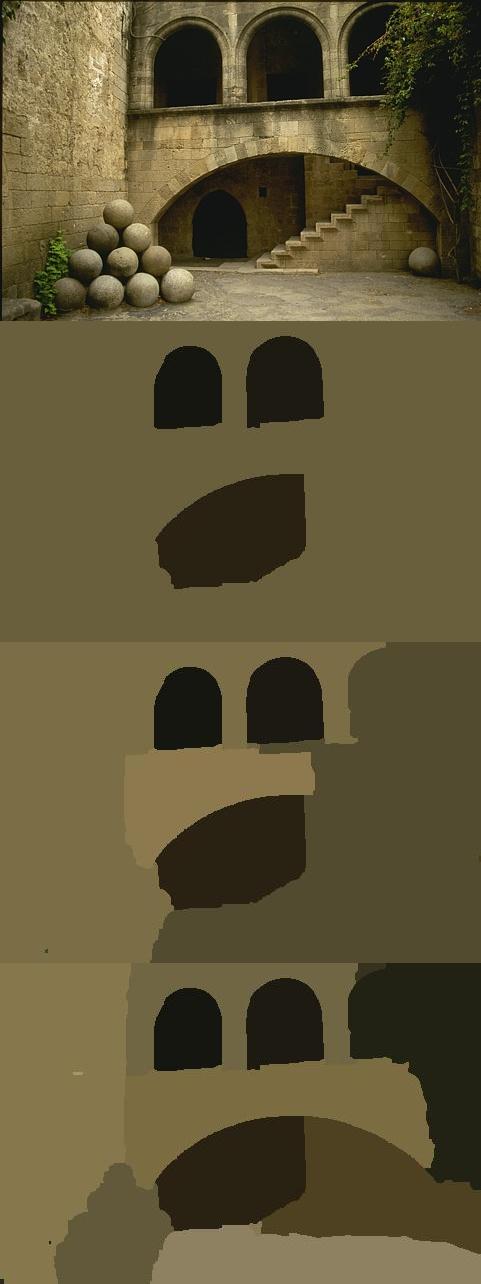}}
\subfigure{\label{quantprb:a}\includegraphics[width=10mm]{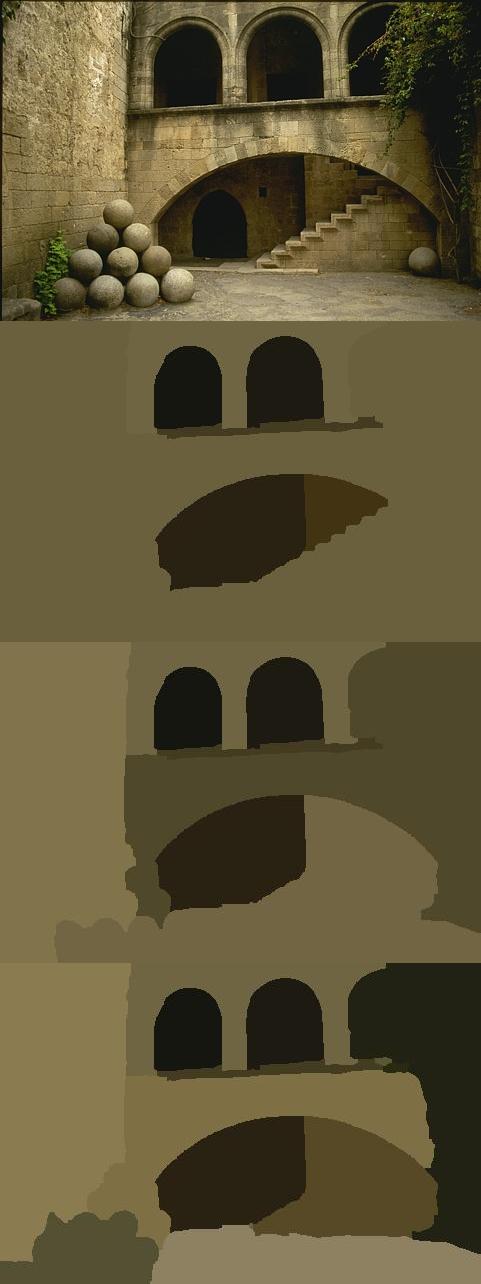}}
\caption{ \textbf{Top}: We display the relative speed of convergence of our two novel algorithms.  We show as a function of time the proportion of problem instances that are not solved up to GAP =$2^{-3,-5,-7}$ from left to right.   (Blue) represents Alg 1 while (red) represents Alg 2.  Timing experiments are ongoing but thousands of data points define these plots.  \textbf{Bottom}:  We show qualitative outputs of Alg 1.  Each group corresponds to a particular image.  We show segmentations of increasing fineness from left to right  with different numbers of random ground truth long range interactions inserted; [28,58,508] from top to bottom .} 
\label{quantprb}
\end{figure*}

Hence we simply compare our two approaches and demonstrate that our algorithms are able to solve the problems in our data set.  We demonstrate the convergence of the  upper bound as a function of time in Fig \ref{quantprb}.  For each algorithm we produce as a function of time the proportion of the problems that algorithm has solved up to a satisfactory level.  This level is computed as follows.  We first compute the maximum lower bound between the two algorithms for a given problem which we denote as $LB$.  We denote the least upper bound produced as a function of time as   $UB(t)$ .  We create a measure called the normalized gap which we define as $GAP(t)=\frac{(UB(t)-LB)}{|LB|}$.  We show rapid convergence and that widest path computation helps improve performance.   


We now study some qualitative results from Alg 1 on various problems.  
For any given problem instance we produce the lowest cost multicut during optimization.  Next for each component in the multicut we compute the mean color of the pixels in that component and color each pixel in the component with that color.   Results are displayed in Fig \ref{quantprb}.  We observe large qualitative improvements in the results.  However when smaller numbers of pairs are used we also often see tiny components created.  However these often disappear and better boundaries emerge as more pairs are used.  


{\small
\bibliographystyle{ieee}
\bibliography{my_bib}
}
\appendix

\section{Computing Any Time Lower Bounds}
\label{proofbound}

In this section we construct a lower bound on the value of the optimal integer multicut that obeys all of the path inequalities. This lower bound is a function of $\lambda$ and $\psi$.  Let $KMCUT$ denote the set of integer multicuts that obey all path inequalities.  We use $X^k$ to denote a member of this set.   
We write $KMCUT$ formally below, along with the optimization over $KMCUT$.  

 \begin{align}
KMCUT=\{ X^k \in MCUT:  SX^k \geq 1 \}\\
\min_{X^k \in KMCUT}\theta X^k
\label{objkmcut}
\end{align}


We now insert Lagrange multipliers corresponding to the constraints in the primal objective in Eq \ref{DualLpFinProperA}.  

\begin{align}
\label{lowero}
\mbox{Eq } \ref{objkmcut} \; =
\min_{X^k \in KMCUT}\max_{\substack{\psi \geq 0\\ \lambda \geq 0} }\theta X^k
+\lambda(X^k-1) 
+\psi(1- SX^k)
\end{align}

Observe that since $X^k$ must lie in KMCUT these multipliers have their global optimum at zero value.  Let any particular setting of $\psi,\lambda$ be chosen and denoted $\hat{\psi}$,$\hat{\lambda}$.   Notice that this produces a lower bound on the true objective  written below.  

\begin{align}
\label{lowereq1}
\mbox{Eq }\ref{lowero} \geq
\min_{X^k \in KMCUT}\theta X^k
+\hat{\lambda}(X^k-1) 
+\hat{\psi}(1- SX^k)
\end{align}
We now relax the constraint that the solution lie in $KMCUT$ to instead lie in the expanded space $MCUT$. 

\begin{align}
\mbox{Eq } \ref{lowereq1} \; \geq
\nonumber \min_{\bar{X} \in MCUT}\theta \bar{X}
+\hat{\lambda}(\bar{X}-1) 
+\hat{\psi}(1- \bar{X})\\
\nonumber =
\min_{\bar{X} \in MCUT}-\lambda 1+\hat{\psi}1+
(\theta+\hat{\lambda}- \psi S)\bar{X}\\
=
-\hat{\lambda} 1+\hat{\psi}1+\min_{\bar{X} \in MCUT}
(\theta+\hat{\lambda}- \psi S)\bar{X}
\label{Eqlower2}
\end{align}

Solving for $\min_{\bar{X}  \in MCUT}(\theta+\hat{\lambda}- \psi S)$ is exactly Planar Correlation Clustering which is known to be NP hard \cite{nphardplanar}.  However \cite{PlanarCC} establishes that the $\frac{3}{2}$ times the value of the minimum 2-colorable multicut lower bounds the value of the optimal multicut.  We use this fact to construct a lower bound on Eq \ref{Eqlower2} below.  

\begin{align}
\mbox{Eq } \ref{Eqlower2} 
\geq -\hat{\lambda} 1+\hat{\psi}1+\frac{3}{2} \min_{z \in C_2}
(\theta+\hat{\lambda}- \psi S)z
\end{align}

Since computing the minimum 2-colorable multicut is solvable in $O(N^{3/2}\log N)$ we can produce a lower bound on the true objective given any setting of $\hat{\lambda}$ and $\hat{\psi}$ which is written below.
\begin{align}
\mbox{Eq } \ref{objkmcut}\geq -\hat{\lambda} 1+\hat{\psi}1+\frac{3}{2} \min_{z \in C_2}
(\theta+\hat{\lambda}- \psi S)z \quad \quad \forall \; [\hat{\lambda} \geq 0,\hat{\psi}\geq 0]
\end{align}

Observe that at convergence of Alg 1,2 the value of $\min_{z\in C_2}(\theta+\hat{\lambda}- \psi S)z=0$ and hence the value of the lower bound is exactly equal to that of the LP relaxation in Eq \ref{DualLpFinProperA}.  During Alg1,2 we are able to produce a lower bound on our objective each time the dual LP solver produces a new setting of $\lambda,\psi$.

\section{Proof Cut}
\label{proofcut}
We now establish that at termination of Alg 1,2 $\min(1,\hat{Z}\gamma +\kappa)$ lies in $CYC$.  First observe that given $\hat{Z}\gamma$  from Alg 1,2 we can greedily decrease values  in $\kappa,\beta$ until each $\kappa_e$ and $\beta_e$ is a term in a tight constraint or is zero valued.  This operation only effects terms associated with zero value in the primal objective.  

We use these reduced $\kappa$ and $\beta$ terms in the remainder of the proof.  We call  property that each  $\kappa$ and $\beta$ term is involved in a tight constraint or is zero valued $Prop1$.

To establish that $\min(1,\hat{Z}\gamma+\kappa) \in CYC$ we  must establish the following for all cycles $c$ and edges $\hat{e}$ in $c$.

\begin{align}
\sum_{e\in c -\hat{e}} \min(1,(\hat{Z}\gamma)_e +\kappa_e) \geq \min(1,(\hat{Z}\gamma)_e +\kappa_e)
\end{align}

On any cycle inequality we refer to the edge on the right side of the inequality as the pivot edge. For short hand we use $Q$ to refer to the set of all pairs of cycle and edge contained in the cycle.  A member of $Q$ is written as $[c,\hat{e}]$ which denotes the cycle and pivot edge.  We associate $Q$ with non-overlapping subsets $Q^+$ and $Q^0$ whose union is $Q$.  An element of $Q$ denoted $[c,\hat{e}]$ is in $Q^+$ if and only if $\kappa_{\hat{e}}>0$.  

We now establish that all cycle inequalities are obeyed by first considering an cycle inequalities in $Q^0$ and then in $Q^+$.

\subsection{For $[c,\hat{e}] \in Q^0$}
To establish that the cycle inequality associated with every pair $[c,\hat{e}] \in Q^0$ we use proof by contradiction.  Consider a violated cycle inequality over pair $[c,\hat{e}]\in Q^0$.  We write this below.  

\begin{align}
\label{eqrt0}
\sum_{e \in c-\hat{e}}\min(1,(\hat{Z}\gamma)_e+\kappa_e)< \min(1,(\hat{Z}\gamma)_{\hat{e}}+\kappa_{\hat{e}})
\end{align}
Since $[c,\hat{e}] \in Q^0$ then $\kappa_{\hat{e}}=0$.  Therefore we write Eq \ref{eqrt0} without $\kappa_{\hat{e}}$.  

\begin{align}
\label{eqrt0a}
\sum_{e \in c-\hat{e}}\min(1,(\hat{Z}\gamma)_e+\kappa_e)< \min(1,(\hat{Z}\gamma)_{\hat{e}})
\end{align}

Recall that \cite{HPlanarCC}  establishes that $\min(1,\hat{Z}\gamma)  \in CYC$ for all $\gamma$ thus the following is true.  

\begin{align}
\label{eqrt}
\sum_{e \in c-\hat{e}}\min(1,(\hat{Z}\gamma)_e)\geq \min(1,(\hat{Z}\gamma)_{\hat{e}})
\end{align}

Since $\kappa$ is non-negative then the following is true which establishes a contradiction with Eq \ref{eqrt0a}.  

\begin{align}
\sum_{e \in c-\hat{e}}\min(1,(\hat{Z}\gamma)_e+\kappa_e)\geq \sum_{e \in c-\hat{e}} \min(1,(\hat{Z}\gamma)_e)\geq (\hat{Z}\gamma)_{\hat{e}}
\end{align}

Therefor the cycle inequalities over members of $Q^0$ are satisfied.

\subsection{For $[c,\hat{e}] \in Q^+$}



To establish that the cycle inequality associated with every pair $[c,\hat{e}] \in Q^+$ is satisfied we use proof by contradiction.  Consider a violated cycle inequality over pair $[c,\hat{e}]\in Q^+$.  We write this below.

\begin{align}
\label{tryme2}
\sum_{e\in \hat{c} -\hat{e}} \min(1,(\hat{Z}\gamma)_e +\kappa_e) < \min(1,(\hat{Z}\gamma)_{\hat{e}} +\kappa_{\hat{e}})
\end{align}

Observe that $ (\hat{Z}\gamma)_{\hat{e}} +\kappa_{\hat{e}} \leq1$ otherwise $\kappa_{\hat{e}}$ would cause a violation of $Prop1$.  
We now alter Eq \ref{tryme2}  based on the observation that $ (\hat{Z}\gamma)_{\hat{e}} +\kappa_{\hat{e}}\leq1$.

\begin{align}
\label{plugmein}
\sum_{e\in \hat{c} -\hat{e}} (\hat{Z}\gamma)_e +\kappa_e < (\hat{Z}\gamma)_{\hat{e}} +\kappa_{\hat{e}}
\end{align}

  Clearly there must be a path inequality including edge $\hat{e}$ that is tight otherwise $PROP1$ is violated.  Denote this path as $p$ and let it connect pair $(f_1,f_2) \in F$.  We now write the expression for that path inequality and alter it.  
\begin{align}
\sum_{e\in p} \min[1,(Z\gamma+\kappa)_e]=1\\
\nonumber=\min[1,(Z\gamma+\kappa)_{\hat{e}}]+\sum_{e\in p-\hat{e}} \min[1,(Z\gamma+\kappa)_e]=1\\
\nonumber=(Z\gamma)_{\hat{e}}+\kappa_{\hat{e}}+\sum_{e\in p-\hat{e}} \min[1,(Z\gamma+\kappa)_e]=1\\
=(Z\gamma)_{\hat{e}}+\kappa_{\hat{e}}+\sum_{e\in p-\hat{e}} (Z\gamma)_e+\kappa_e=1
\label{botgroup}
\end{align}

We now substitute the left hand side of Eq \ref{plugmein} for $(\hat{Z}\gamma)_{\hat{e}} +\kappa_{\hat{e}}$ in Eq \ref{botgroup}.  

\begin{align}
1=(Z\gamma)_{\hat{e}}+\kappa_{\hat{e}}+\sum_{e\in p-\hat{e}} (Z\gamma)_e+\kappa_e> \sum_{e\in \hat{c} -\hat{e}} (\hat{Z}\gamma)_e +\kappa_e +\sum_{e\in p-\hat{e}} (Z\gamma)_e+\kappa_e
\label{mustbeneg}
\end{align}
Observe that the union of $\hat{c}$ and $p-\hat{e}$ is itself a path between a pair $f_1,f_2 \in F$ though one which may include edges multiple times.  We denote this path as $\bar{p}$.  Now observe the following. 
\begin{align}
1=(Z\gamma)_{\hat{e}}+\kappa_{\hat{e}}+\sum_{e\in p-\hat{e}} (Z\gamma)_e+\kappa_e\\
\nonumber > \sum_{e\in \hat{c} -\hat{e}} (\hat{Z}\gamma)_e +\kappa_e +\sum_{e\in p-\hat{e}} (Z\gamma)_e+\kappa_e\\
\nonumber \geq \sum_{e \in {\bar{p}}}(Z\gamma)_e+\kappa_e
\label{mustbenegdd}
\end{align}
However since path $\bar{p}$ connects a pair $f_1,f_2 \in F$ then the sum of the edges on the path must be greater than or equal to one.  Thus we have established a contradiction and therefore the pair  $\hat{e},c$ is not associated with a violated cycle inequality.

\subsection{Conclusion}

Since every pair $[c,\hat{e}] \in Q^+ \cup Q^0$ the associated cycle inequality is satisfied then all cycle inequalities are satisfied and thus $\min(1,\hat{Z}\gamma +\kappa) \in CYC$.

\end{document}